\documentclass{article}

    \PassOptionsToPackage{numbers, compress}{natbib}

\usepackage[preprint]{neurips_2024}

\usepackage[utf8]{inputenc} 
\usepackage[T1]{fontenc}    
\usepackage{hyperref}       
\usepackage{url}            
\usepackage{booktabs}       
\usepackage{amsfonts}       
\usepackage{nicefrac}       
\usepackage{microtype}      
\usepackage{xcolor}         
\usepackage{amsthm}
\usepackage{xspace}
\usepackage{enumitem}
\usepackage{adjustbox}
\usepackage{makecell}
\usepackage{multirow} 
\usepackage{mathtools}

\usepackage{amsthm}
\usepackage{color}

\newcommand{\cut}[1]{{}}

\newcommand{\tA}{{\tilde{\vA}}}

\newcommand{\tL}{{\tilde{\vL}}}

\newcommand{\vb}{{\mathbf{b}}}

\newcommand{\vw}{{\mathbf{w}}}

\newcommand{\vy}{{\mathbf{y}}}

\newcommand{\vA}{{\mathbf{A}}}

\newcommand{\vD}{{\mathbf{D}}}

\newcommand{\vI}{{\mathbf{I}}}

\newcommand{\vL}{{\mathbf{L}}}

\newcommand{\vU}{{\mathbf{U}}}

\newcommand{\vX}{{\mathbf{X}}}
\newcommand{\vY}{{\mathbf{Y}}}

\newcommand{\vLambda}{{\mathbf{\Lambda}}}

\newcommand{\RR}{\mathbb{R}}

\newcommand{\bmu}{\boldsymbol{\mu}}
\newcommand{\bnu}{\boldsymbol{\nu}}

\makeatletter
\let\@@span\span
\def\sp@n{\@@span\omit\advance\@multicnt\m@ne}
\makeatother

\newcommand{\bc}{\begin{center}}
\newcommand{\ec}{\end{center}}

\newcommand{\bdm}{\begin{displaymath}}
\newcommand{\edm}{\end{displaymath}}

\newcommand{\beq}{\begin{equation}}
\newcommand{\eeq}{\end{equation}}

\newcommand{\bfl}{\begin{flushleft}}
\newcommand{\efl}{\end{flushleft}}

\newcommand{\bt}{\begin{tabbing}}
\newcommand{\et}{\end{tabbing}}

\newcommand{\beqn}{\begin{align}}
\newcommand{\eeqn}{\end{align}}

\newcommand{\beqs}{\begin{align*}} 
\newcommand{\eeqs}{\end{align*}}  

\newtheorem{theorem}{Theorem}

\newtheorem{definition}{Definition}

\newcommand{\method}{\textsc{Node-MoE}\xspace }

\title{Node-wise Filtering in Graph Neural Networks: \\ A Mixture of Experts Approach}

\author{%
  Haoyu Han \\
  Michigan State University\\
  \texttt{hanhaoy1@msu.edu} \\
   \And
   Juanhui Li \\
   Michigan State University \\
   \texttt{lijuanh1@msu.edu} \\
   \And
   Wei Huang \\
   RIKEN AIP \\
   \texttt{wei.huang.vr@riken.jp} \\
   \And
   Xianfeng Tang \\
   Amazon \\
   \texttt{xianft@amazon.com} \\
   \And
   Hanqing Lu \\
   Amazon \\
   \texttt{luhanqin@amazon.com} \\
    \And
   Chen Luo \\
   Amazon \\
   \texttt{cheluo@amazon.com} \\
   \And
      Hui Liu \\
   Michigan State University \\
   \texttt{liuhui7@msu.edu} \\
   \And
      Jiliang Tang \\
   Michigan State University \\
   \texttt{tangjili@msu.edu} 
}

\begin{document}

\maketitle

\begin{abstract}
Graph Neural Networks (GNNs) have proven to be highly effective for node classification tasks across diverse graph structural patterns. Traditionally, GNNs employ a uniform global filter—typically a low-pass filter for homophilic graphs and a high-pass filter for heterophilic graphs. However, real-world graphs often exhibit a complex mix of homophilic and heterophilic patterns, rendering a single global filter approach suboptimal.  In this work, we theoretically demonstrate that a global filter optimized for one pattern can adversely affect performance on nodes with differing patterns. To address this, we introduce a novel GNN framework  \method that utilizes a mixture of experts to adaptively select the appropriate filters for different nodes.  Extensive experiments demonstrate the effectiveness of \method on both homophilic and heterophilic graphs.
\end{abstract}

\section{Introduction}
\label{sec:intro}




Graph Neural Networks (GNNs)~\citep{kipf2016semi, velivckovic2017graph} have emerged as powerful tools in representation learning for graph structure data, and have achieved remarkable success on various graph learning tasks~\citep{wu2020comprehensive, ma2021deep}, especially the node classification task. GNNs usually can be designed and viewed from two domains, i.e., spatial domain and spectral domain. In the spatial domain, GNNs~\citep{kipf2016semi, hamilton2017inductive, gasteiger2018predict} typically follow the message passing mechanism~\citep{gilmer2017neural}, which propagate messages between neighboring nodes. In the spectral domain, GNNs~\citep{defferrard2016convolutional, chien2020adaptive} apply different filters on the graph signals in the spectral domain of the graph Laplacian matrix. 

Most GNNs have shown great effectiveness in the node classification task of homophilic graphs~\citep{velivckovic2017graph, wu2019simplifying, gasteiger2018predict, baranwal2021graph}, where connected nodes tend to share the same labels. These GNNs usually leverage the low-pass filters, where the smoothed signals are preserved. However, the heterophilic graphs exhibit the heterophilic patterns, where the connected nodes tend to have different labels. As a result, several GNNs~\citep{sun2022improving, li2024pc, bo2021beyond} designed for heterophilic graphs introduce the high-pass filter to better handle such diversity. To adapt to both homophilic and heterophilic graphs, GNNs with learnable graph convolution~\cite{chien2020adaptive, bianchi2021graph, he2021bernnet, he2022convolutional} can automatically learn different types of filters for different types of graphs. Despite the great success, these GNNs usually apply a uniform global filter across all nodes.

\begin{figure*}[htb]
    \centering
   \includegraphics[width=1\linewidth]{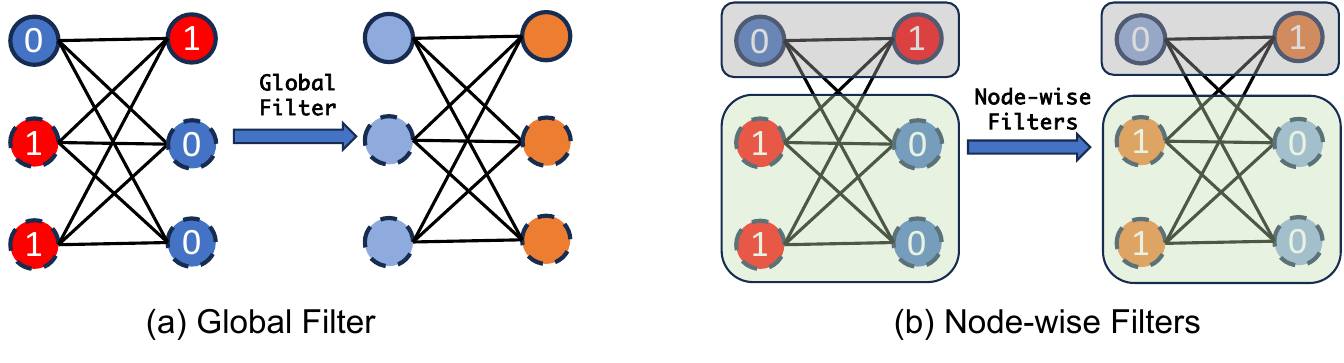}
    \caption{A toy example to illustrate the effect of global and node-wise filters. The node color represents features, and the number indicates the labels.}
    \label{fig:example}
\end{figure*}

However, real-world graphs often display a complex interplay of homophilic and heterophilic patterns~\citep{li2022finding, mao2024demystifying}, challenging this one-size-fits-all filtering approach. Specifically, while some nodes tend to connect with others that share similar labels, reflecting homophilic patterns, others are more inclined to form connections with nodes that have differing labels, indicative of heterophilic patterns.
Applying a uniform type of filter, tailored for just one of these patterns, across all nodes may hurt the performance of other patterns. To illustrate this, we provide an example as shown in Figure~\ref{fig:example}(a), where different colors represent distinct node features, and numbers indicate node labels. The nodes are marked as either solid or dotted circles to denote homophilic and heterophilic patterns, respectively. Applying a global low-pass filter, such as the adjacency matrix $\vA$, uniformly across all nodes results in a scenario where nodes on the left possess the same feature, while those on the right possess another. Therefore, all the left nodes or the right nodes will have the same prediction. However, nodes on the left or right don't share the same label. 
Consequently, this global filtering approach leads to misclassification of nodes.

This toy example clearly illustrates the limitations of a one-size-fits-all filtering strategy and motivates the need for a more tailored approach. To address this, we propose applying different filters to nodes based on their specific structural patterns. Figure~\ref{fig:example}(b) provides an example that we apply a low-pass filter, such as $\vA$, to homophilic nodes, and a high-pass filter, such as $-\vA$, to heterophilic nodes. From the results, nodes in the same class would have the same features. Therefore, this node-wise filtering approach allows for the perfect classification of all nodes in this example.

\textbf{Present work.} In this work, we observe that nodes in many real-world graphs not only exhibit diverse structural patterns but also that these patterns vary significantly among different communities within the same graph.
Utilizing the CSBM model to generate graphs with mixed structural patterns, we theoretically demonstrate that a global filter optimized for one pattern may incur significant losses for nodes with other patterns, while node-wise filtering can achieve linear separability for all nodes under mild conditions. Building on these insights, we propose a Node-wise filtering method - \method, which leverages a Mixture of Experts framework to adaptively select appropriate filters for different nodes. Extensive experiments validate the effectiveness of the proposed \method on both homophilic and heterophilic graphs, illustrating significant performance improvement.

\section{Preliminary}
\label{sec:pre}

In this section, we explore the structural patterns present in various graph datasets, which usually exhibit mixed homophilic and heterophilic patterns. Then, we theoretically demonstrate that a global filter often fails in graphs characterized by such mixed structural patterns. In contrast, node-wise filtering can achieve linear separability under mild conditions. Before we start, we first define the notations used in this paper and background knowledge.

\textbf{Notations.} We use bold upper-case letters such as $\mathbf{X}$ to denote matrices. $\mathbf{X}_i$ denotes its $i$-th row and $\vX_{ij}$ indicates the $i$-th row and $j$-th column element. We use bold lower-case letters such as $\mathbf{x}$ to denote vectors. Let $\mathcal{G}=(\mathcal{V}, \mathcal{E})$ be a graph, where $\mathcal{V}$ is the node set, $\mathcal{E}$ is the edge set, and $|\mathcal{V}|=n$. 
$\mathcal{N}_i$ denotes the neighborhood node set for node $v_i$.  
The graph can be represented by an adjacency matrix $\mathbf{A} \in \mathbb{R}^{n \times n}$, where $\mathbf{A}_{ij}>0$ indices that there exists an edge between nodes $v_i$ and $v_j$ in $\mathcal{G}$, or otherwise $\mathbf{A}_{ij}=0$. For a node $v_i$, we use $\mathcal{N}(v_i)=\{v_j: \vA_{ij} > 0\}$ to denote its neighbors. Let $\vD = diag(d_1, d_2, \dots, d_n)$ be the degree matrix, where $d_i = \sum_{j}\mathbf{A}_{ij}$ is the degree of node $v_i$. Furthermore, suppose that each node is associated with a $d$-dimensional feature $\mathbf{x}$ and we use $\mathbf{X} = [ \mathbf{x}_1, \dots, \mathbf{x}_n ]^{\top} \in \mathrm{R}^{n \times d}$ to denote the feature matrix. Besides, the label matrix is $\vY \in \mathrm{R}^{n \times c}$, where $c$ is the number of classes. We use $y_u$ to denote the label of node $u$.

\textbf{Graph Laplacian.} The graph Laplacian matrix is defined as $\mathbf{L} = \mathbf{D} - \mathbf{A}$. We define the normalized adjacency matrix as $\tA = {\mathbf{D}}^{-\frac{1}{2}} {\mathbf{A}} {\mathbf{D}}^{-\frac{1}{2}}$ and the normalized Laplacian matrix as $\tL = \vI - \tA$. Its eigendecomposition can be represented by $\tL=\vU\vLambda\vU^\top$, where the $\vU \in \mathbb{R}^{n \times n}$ is the eigenvector matrix and $\vLambda=diag(\lambda_1, \lambda_2, \dots, \lambda_n)$ is the eigenvalue matrix. Specifically, $0 \leq \lambda_1 \leq \lambda_2 \leq \dots \leq \lambda_n < 2$. The filtered signals can be represented by $\hat{\vX} = \vU f(\vLambda)\vU^\top\vX$, where $f$ is the filter function. As a result, the graph convolution $\tA\vX$ can be viewed as a low-pass filter, with the filter $f(\lambda_i) = 1-\lambda_i$. Similarly, the graph convolution $-\tA\vX$ is a high-pass filter with filter $f(\lambda_i) = \lambda_i - 1$. 

\textbf{Homophily metrics.} Homophily metrics in graphs measure the tendency of edges to connect nodes with similar labels~\citep{platonov2024characterizing}. There are several commonly used homophily metrics, such as edge homophily~\citep{zhu2020beyond}, node homophily~\citep{pei2020geom}, and class homophily~\citep{lim2021new}. In this paper, we adopt the node homophily $H(\mathcal{G}) = \frac{1}{|\mathcal{V}|}\sum_{v_i \in \mathcal{V}}h(v_i)$, where $h(v_i) = \frac{\left|\left\{u \in \mathcal{N}\left(v_i\right): y_u=y_v\right\}\right|}{d_i}$ measures the label similarity between node $v_i$ with its neighbors. A node with higher $h(v)$ exhibits a homophilic pattern while a low $h(v)$ indicates a heterophilic pattern.

\subsection{Structural Patterns in Existing Graphs}
\label{sec:pattern}
In this subsection, we examine the structural patterns present in existing graph datasets. Specifically, we select two widely used homophilic datasets, i.e., Cora and CiteSeer~\citep{sen2008collective}, and two heterophilic datasets, i.e., chameleon and squirrel~\cite{rozemberczki2021multi}.
\begin{figure*}[htb]
\centering
\vspace{-0.1in}
\begin{minipage}[b]{.48\textwidth}
  \centering
  \includegraphics[width=\linewidth]{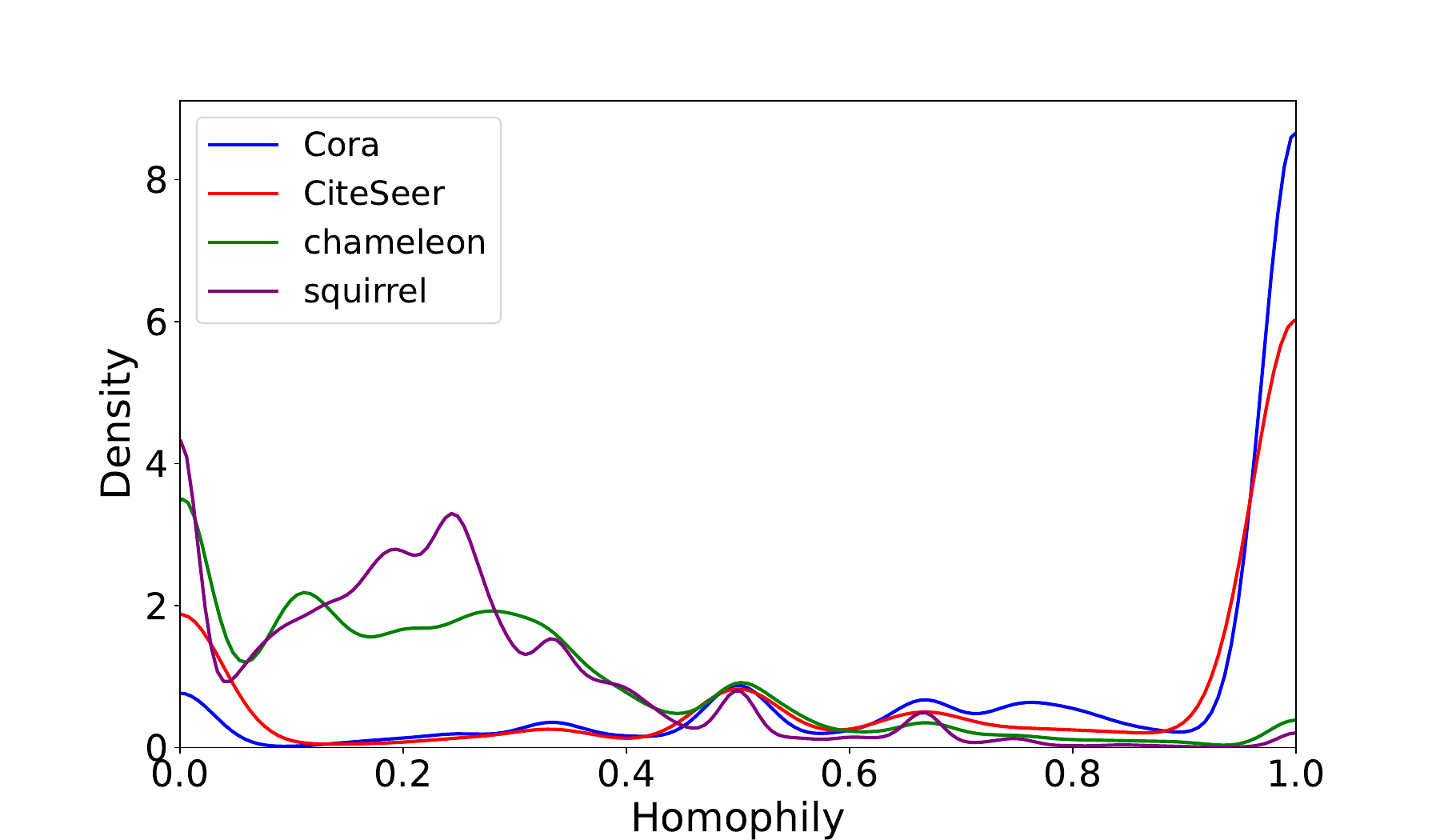}
\caption{Node homophily density.}
\label{fig:homophily_density}
\end{minipage} \hfill
\begin{minipage}[b]{.48\textwidth}
  \centering
  \includegraphics[width=\linewidth]{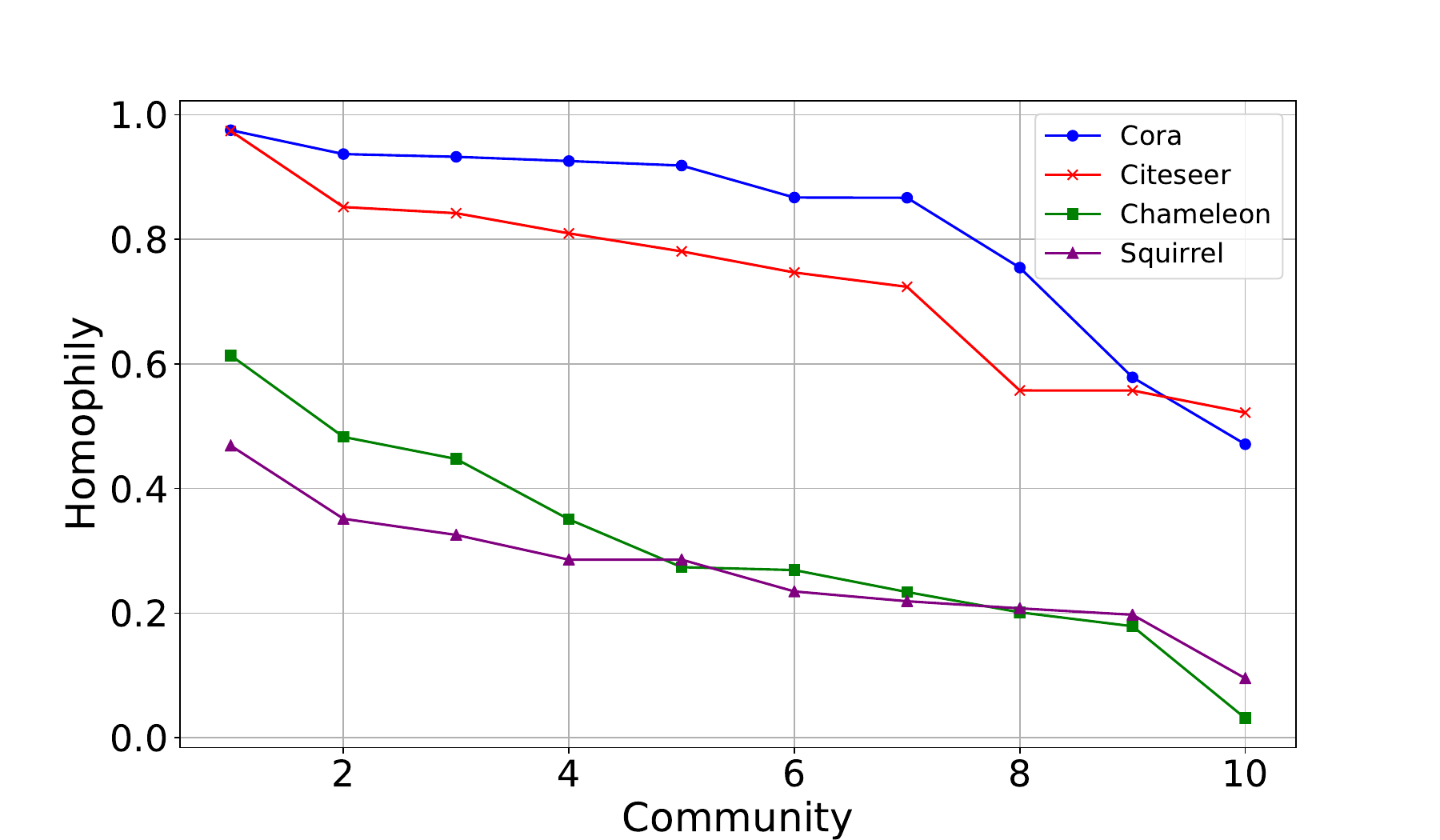}
    \caption{Homophily in different communities.}
 \label{fig:community}
\end{minipage}
\vspace{-0.1in}
\end{figure*}
We first calculate the homophily distribution for all nodes in the graph. As shown in Figure~\ref{fig:homophily_density}, while the majority of nodes in homophilic graphs predominantly exhibit homophilic patterns, and those in heterophilic graphs display heterophilic patterns, exceptions are evident. Notably, some nodes in homophilic graphs show heterophilic tendencies, and conversely, some nodes in heterophilic graphs demonstrate homophilic patterns. Consequently, \textbf{all these graphs exhibit a mixture of homophilic and heterophilic patterns}, which aligns with the findings in the previous works~\cite{li2022finding, mao2024demystifying}.

We further analyze the position of nodes with different structural patterns within the graphs. To do this, we divide each graph into several subgraphs using community detection algorithms~\cite{fortunato2010community}. We focus on the largest 10 communities and calculate the homophily level for each subgraph. The results, as shown in Figure~\ref{fig:community}, reveal \textbf{significant variations in homophily across different communities}. For instance, in the Cora dataset, homophily levels in some communities approach 1, indicating strong homophily, while in some communities it drops below 0.5. Similarly, in the chameleon dataset, the lowest homophily levels are near 0, with the highest reaching above 0.6. These findings highlight the considerable diversity in node interaction patterns, even within the same graph, underscoring the complexity of graph structures in real-world datasets. The variability in homophily levels clearly illustrates that nodes in various parts of the graph may require distinct processing approaches. Therefore, applying the same global filter to all nodes may lead to suboptimal performance.

\subsection{Analysis based on CSBM model}
\label{sec:analysis}
To further illustrate why applying a global filter may result in suboptimal performance, we utilize the Contextual Stochastic Block Model (CSBM)~\citep{deshpande2018contextual}, which has been widely applied to graph analysis~\cite{fortunato2016community, jiang2023topology}, such as analyzing the behavior of GNNs~\citep{palowitch2022synthetic, baranwal2021graph, ma2021homophily}. The CSBM is a generative model, which is often used to generate graph structures and node features. Typically, CSBMs are based on the assumption that graphs are generated following a uniform patter, such as nodes with the same label are connected with probability $p$ while nodes with different labels are connected with probability $q$~\cite{ma2021homophily}. However, the real-world complexity of graphs features a mixture of homophilic and heterophilic patterns, as illustrated in section~\ref{sec:pattern}. We adapt the CSBM by mixing two CSBMs to generate one graph, mirroring the approach~\cite{mao2024demystifying}.

\begin{definition}
\label{def1}
{$CSBM(n, \bmu, \bnu, (p_0, q_0), (p_1, q_1), P)$. The generated nodes consist of two classes, $C_0=\left\{i \in[n]: y_i=0\right\}$ and $C_1=\left\{j \in[n]: y_j=1\right\}$. For each node, consider $\vX \in \RR^{n \times d}$ to be the feature matrix such that each row $\vX_i$ is an independent $d$-dimensional Gaussian random vectors with $\vX_i \sim N\left(\boldsymbol{\mu}, \frac{1}{d} \vI\right)$ if $i \in C_0$ and $\vX_j \sim N\left(\boldsymbol{\nu}, \frac{1}{d} \vI\right)$ if $j \in C_1$. Here $\boldsymbol{\mu}, \boldsymbol{\nu}$ are the fixed class mean vector with $\|\boldsymbol{\mu}\|_2,\|\boldsymbol{\nu}\|_2 \leq 1$ and $\vI$ is the identity matrix. Suppose there are two patterns of nodes in the adjacency matrix $\vA=(a_{ij})$, i.e., the homophilic pattern:  $H_0 = \{i \in[n]: a_{ij} = \operatorname{Ber}(p_0)$ \text{if} $y_i = y_j $ and
$a_{ij} = \operatorname{Ber}(q_0)$ if $y_i \neq y_j , p_0 > q_0 \}$ and the heterophilic pattern: $H_1 = \{i \in[n]: a_{ij} = \operatorname{Ber}(p_1)$ \text{if} $y_i = y_j $ and
$a_{ij} = \operatorname{Ber}(q_1)$ if $y_i \neq y_j , p_1 < q_1 \}$. $P$ denotes the probability that a node is in the homophilic pattern. We also assume the nodes follow the same degree distribution $p_0+q_0=p_1+q_1$}.
\end{definition}

For simplification, we consider a linear model with parameters $\vw \in \mathbb{R}^d$ and $b \in \mathbb{R}$, following the approach~\cite{baranwal2021graph}. The predicted label for nodes is given by $\hat{\vy} = \sigma(\tilde{\vX}\vw + b\mathbf{1})$, where $\sigma(x) = (1+e^{-x})^{-1}$ is the sigmoid function, and $\tilde{\vX}$ represents the features after filtering.  The binary cross-entropy loss over nodes $\mathcal{V}$ is formulated as $L(\mathcal{V}, \vw, b) = -\frac{1}{|\mathcal{V}|}\sum_{i \in \mathcal{V}} y_i \log(\hat{y_i}) + (1-y_i)\log(1-\hat{y_i})$.

\begin{theorem}
\label{theorem1}
Suppose $n$ is relatively large, the graph is not too sparse with $p_i, q_i = \omega(\log ^2(n) / n)$ and the feature center distance is not too small with $\|\boldsymbol{\mu}-\boldsymbol{\nu}\|=\omega(\frac{\log n}{\sqrt{\operatorname{dn}(p_0+q_0)}})$ and $\|\vw\| \leq R$. For the graph $G(\mathcal{V}, \mathcal{E}, \vX) \sim CSBM(n, \bmu, \bnu, (p_0, q_0), (p_1, q_1), P)$, we have the following:

1. If the low-pass global filter, i.e., $1-\lambda$, is applied to the whole graph $G$, we can find a optimal $\vw^{*}, b^*$ that achieve near linear separability for the homophilic node set $H_0$. However, the loss for the heterophilic node set $H_1$ can be relatively large with:
\begin{align*}
    L(H_1, \vw^{*}, b^*) \geq \frac{R(q_1-p_1)}{2(q_1+p_1)}\|\boldsymbol{\mu}-\boldsymbol{\nu}\|\left(1+o_d(1)\right).
\end{align*}

2. If different filters are applied to homophilic and heterophilic sets separately,  we can find an optimal $\vw^{*}, b^*$ that all the nodes are linear separable with the probability:
\begin{align*}
    \mathbb{P}\left(\left(\tilde{\vX}_i\right)_{i \in \mathcal{V}} \text { is linearly separable }\right)=1-o_d(1).
\end{align*}
\end{theorem}

The proof of these results is detailed in Appendix~\ref{app:proof}. Theorem~\ref{theorem1} reveals critical insights into the filtering strategies for graphs with mixed homophilic and heterophilic patterns, as generated by the CSBM model. The first part of the theorem illustrates that applying a global low-pass filter can create an optimal classifier for homophilic nodes, achieving near-linear separability. However, this classifier may result in a substantial loss for heterophilic nodes, highlighting the limitations of a uniform filtering strategy. Conversely, the second part of the theorem demonstrates that by applying different filters to different patterns of nodes separately, it is possible to achieve linear separability across all nodes. These findings strongly motivate the exploration of a node-wise filtering method, which can automatically apply different filters to distinct nodes based on their specific patterns, to improve the overall performance.

\section{The Proposed Method}
\label{sec:method}
The investigations presented in Section~\ref{sec:pre} underscore the complex nature of real-world datasets, revealing a mixture of homophilic and heterophilic patterns within them. Additionally, these patterns are not uniformly distributed throughout the graph; rather, the level of homophily varies significantly across different communities.  Our theoretical analysis further demonstrates that global filtering, as commonly employed in numerous GNNs, may not effectively capture such complex patterns, often leading to suboptimal performance. In contrast, node-wise filtering, which applies distinct filters to individual nodes based on their specific patterns, shows great promise in handling the intricacies of such complex graphs. 

However, implementing the node-wise filtering approach presents two significant challenges. First, how can we incorporate various filters into a single unified framework? It requires a flexible architecture that can seamlessly accommodate multiple filtering mechanisms without compromising the efficiency and scalability of the model. Second, without ground truth on node patterns, how can we select the appropriate filters for different nodes? In the following subsections, we aim to address these challenges.

\begin{figure*}[htb]
    \centering
   \includegraphics[width=1\linewidth]{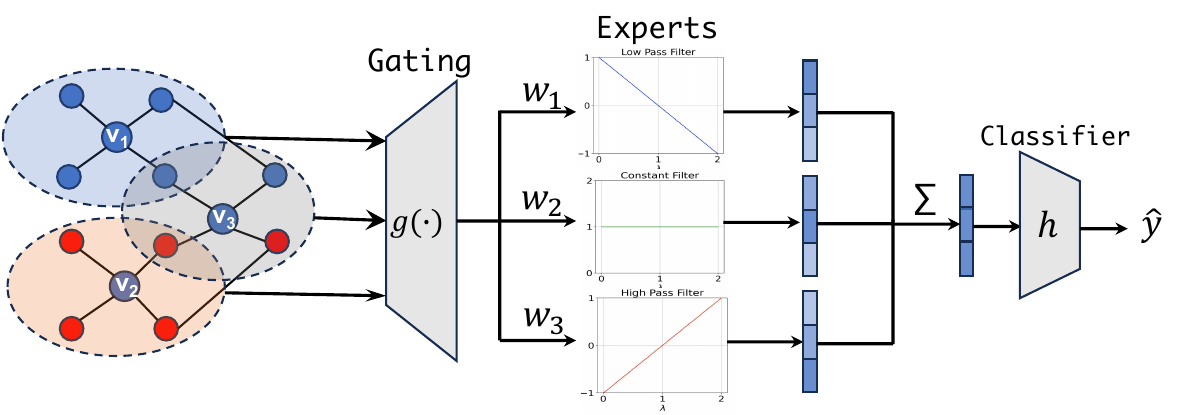}
    \caption{The overall framework of the proposed \method. For each node, the gating model will assign different weights for each expert based on the node's feature and context. The experts can be any GNNs with different filters.}
    \label{fig:framework}
\end{figure*}

\subsection{\method: Node-wise Filtering via Mixture of Experts}
Mixture of Experts (MoE)~\citep{jacobs1991adaptive, jordan1994hierarchical}, which follows the divide-and-conquer principle to divide the complex problem space into several subspaces so that each one can be easily addressed by specialized experts, have been successfully adopted across various domains~\cite{masoudnia2014mixture, shazeer2017outrageously, riquelme2021scaling}. For node classification tasks in graphs exhibiting a mixture of structural patterns, the diversity of node interactions necessitates applying distinct filters to different nodes as we discussed in Sections~\ref{sec:pre}. This necessity aligns well with the MoE methodology, which processes different samples with specific experts. Building on this principle, we introduce a flexible and efficient Node-wise Filtering via Mixture of Experts (\method) framework, designed to dynamically apply appropriate filters to nodes based on their structural characteristics.

The overall \method framework is illustrated in Figure~\ref{fig:framework}, which consists of two primary components: the gating model and the multiple expert models. With the graph data as input, the gating model $g(\cdot)$ computes the weight assigned to each expert for every node, reflecting the relevance of each expert's contribution to that specific node. Each expert model, implemented as any GNN with different filters, generates node representations independently. The final node classification is determined by a weighted sum of these representations, where the weights are those assigned by the gating model. The prediction for node $i$ can be represented by:

\begin{align}
    \hat{y}_{i}=h \left(\sum_{o=1}^m g(\vA, \vX)_{i, o} E_o(\mathbf{A}, \mathbf{X})_{i}\right),
\end{align}

where $m$ is the number of experts, $E_o$ denotes the $o$-th expert, $g(\mathbf{A}, \mathbf{X})_{i, o}$ represents the weight assigned to the $o$-th expert for node $i$ by the gating model, and $h$ is a classifier, which could be a model like a neural network or a simple activation function like Softmax. In the following, we will delve into the specific designs of the gating model and the expert models within the proposed Node-MoE framework.

\subsection{Gating Model}

The gating model is a pivotal component of the Node-MoE framework, aimed at selecting the most appropriate experts for each node. Its primary function is to dynamically assign higher weights to experts whose filtering characteristics best match the node's patterns. For instance, an expert utilizing a high-pass filter may receive a higher weight for a node that exhibits heterophilic patterns. However, a significant challenge arises as there is no explicit ground truth indicating which pattern each node belongs to. In traditional MoE models, the gating model often utilizes a straightforward feed-forward network that processes the features of the sample as input~\citep{shazeer2017outrageously, riquelme2021scaling, du2022glam, wang2024graph}. Nevertheless, the nodes with different patterns may share similar node features, making this method ineffective. 

To address this challenge, we estimate node patterns by incorporating the contextual features surrounding each node. If a node's features significantly differ from those of its neighboring nodes, it is likely that this node exhibits a heterophilic pattern. Specifically, the input to our gating model includes a composite vector $[\vX, |\vA\vX - \vX|, |\vA^2\vX-\vX|]$. This vector combines the node's original features with the absolute differences between its features and those of its neighbors over one and two hops, respectively, to indicate the node's structural patterns. Moreover, as discussed in Section~\ref{sec:pattern}, different structural patterns are not uniformly distributed across the graph, and distinct communities may exhibit varying structural characteristics. To capitalize on this phenomenon, we employ GNNs with low-pass filters, such as GIN~\citep{xu2018powerful}, for the gating model.  These networks are chosen due to their strong community detection capabilities~\cite{shchur2019overlapping, bruna2017community}, ensuring that neighboring nodes are likely to receive similar expert selections. Experimental results in Section~\ref{sec:case_study} clearly demonstrate the proposed gating can efficiently assign different nodes to their suitable filters. 

\subsection{Expert Models}
The mixed structural patterns observed in real-world graphs necessitate that the expert models in our \method framework possess diverse capabilities. To achieve this, we consider multiple existing GNNs equipped with different filters. Traditional GNNs often utilize fixed filters, which may not adequately capture the complexity of diverse structural patterns. To address this limitation, we opt for GNNs with learnable graph convolutions~\citep{chien2020adaptive, bianchi2021graph, he2021bernnet, he2022convolutional}, which are capable of adapting their filters to better fit the graph structural patterns. However, the same experts would make the gating model hard to learn the right features~\citep{chen2022towards} and may result in all experts' filters being optimized in the same direction. To encourage diversity and ensure that each expert is adept at handling specific structural patterns, we adopt a differentiated initialization strategy for the filters in the experts. Instead of using a fixed filter initialization, we initialize different experts with distinct types of filters, such as low-pass, constant, and high-pass filters.

\textbf{Filter Smoothing Loss.} While integrating multiple experts with diverse filters significantly enhances the expressive capacity of our \method framework, this complexity can also make the model more challenging to fit. For example, training multiple filters simultaneously may lead to oscillations in the spectral domain for each filter as shown in Appendix~\ref{app:smoothing}. This not only complicates fitting the model to the data but also impacts its explainability. The specific role and function of each oscillating filter become difficult to discern, making it harder to understand and interpret the model's behavior. To mitigate these issues, we introduce a filter smoothing loss designed to ensure that the learned filters exhibit smooth behavior in the spectral domain. This loss is defined as follows:
\begin{align}
    L_s^o = \sum_{i = 1}^K |f_o(x_i) - f_o(x_{i-1})|^2,
\end{align}
where $f_o(\cdot)$ is the learnable filter function of the $o$-th expert,  $x_0 \leq x_1 \leq \cdots \leq x_K$ are $K+1$ values spanning the spectral domain. The overall training loss is then given by  $L = L_{task} + \gamma \sum_{o=1}^m L_s^o$, where the $L_{task}$ is the node classification loss and $\gamma$ is a hyperparameter that adjusts the influence of the filter smoothing loss.

\subsection{Top-K gating}
\label{sec:topk}
The soft gating that integrates all experts in the Node-MoE framework significantly enhances its modeling capabilities, but it also increases computational complexity since each expert must process all samples. To improve computational efficiency while maintaining performance, we introduce a variant of \method by leveraging the Top-K gating mechanism~\citep{shazeer2017outrageously}. In this variant, the \method with Top-K gating selectively activates only the top k experts with the highest relevance for each node.  Specifically, the gating function for a node $v_i$ is defined as $g(v_i)=\operatorname{Softmax}\left(\operatorname{TopK}\left(g\left(\vA, \vX\right)_i, k\right)\right)$. To prevent the gating model from consistently favoring a limited number of experts, we incorporate a load-balancing loss as suggested by \cite{shazeer2017outrageously}. In the experiments detailed in Section~\ref{sec:ablation}, we set $k=1$ to ensure that the computational complexity of our \method is comparable to that of an individual expert model.

\section{Experiment}
\label{sec:exp}
In this section, we conduct comprehensive experiments to validate the effectiveness of the proposed \method. Specifically, we aim to address the following research questions:

\begin{itemize} [leftmargin=0.2in]
    \item \textbf{RQ1:} How does \method perform compared with the state-of-the-art baselines on both homophilic and heterophilic graphs?
    \item \textbf{RQ2:} Do the experts within \method learn diverse structural patterns and does the gating model accurately assign each node to its most suitable experts?
    \item \textbf{RQ3:}  How do different factors affect the performance of \method?
\end{itemize}

\subsection{Experimental settings.}
\label{sec:exp_settings}
\textbf{Datasets.} To evaluate the efficacy of our proposed \method, we conduct experiments across seven widely used datasets. These include four homophilic datasets: Cora, CiteSeer, Pubmed~\citep{sen2008collective}, and ogbn-arxiv~\citep{hu2020open}; along with three heterophilic datasets: Chameleon, Squirrel, and Actor~\citep{pei2020geom}. For Cora, CiteSeer, and Pubmed, we generate ten random splits, distributing nodes into 60\% training, 20\% validation, and 20\% testing partitions.  For the heterophilic datasets, we utilize the ten fixed splits as specified in \cite{pei2020geom}. The ogbn-arxiv dataset is evaluated using its standard split~\citep{hu2020open}. We run the experiments 3 times for each split
and report the average performance and standard deviation. More details about these datasets are shown in Appendix~\ref{sec:app_data}.

\textbf{Baselines.} We compare our method with a diverse set of baselines, which can be divided into five categories: (1) Non-GNN methods like MLP and Label Propagation (LP)~\citep{zhou2003learning}; (2) Homophilic GNNs utilizing fixed low-pass filters such as  GCN~\citep{kipf2016semi}, GAT~\citep{velivckovic2017graph}, APPNP~\citep{gasteiger2018predict}, and GCNII~\citep{chen2020simple}; (3) Heterophilic GNNs including WRGCN~\citep{suresh2021breaking}, GloGNN~\citep{li2022finding} and LinkX~\citep{lim2021large}; (4) GNNs with learnable filters like GPRGNN~\citep{chien2020adaptive} and ChebNetII~\citep{he2022convolutional}; (5) MoE-based GNNs such as GMoE~\citep{wang2024graph}.

\textbf{\method settings.} The proposed \method framework is highly flexible, allowing for a wide range of choices in both gating and expert models. In this work, we employ the GIN~\citep{xu2018powerful} as the gating model due to its exceptional expressive power and ability to leverage community properties as discussed in Section~\ref{sec:pre}. As for the expert models, we utilize ChebNetII~\citep{he2022convolutional}, known for its efficiency in learning filters. Specifically, we experiment with configurations of 2, 3, and 5 ChebNetII experts, each initialized with different filters. More details and parameter settings are in Appendix~\ref{sec:app_expsetting}.

\subsection{Performance Comparison on Benchmark Datasets}
In this section, we evaluate the efficacy of the proposed \method across both homophilic and heterophilic datasets. The results of node classification experiments are detailed in Table~\ref{tab:performance}. From the results, we can have the following observations:

\begin{itemize} [leftmargin=0.2in]
    \item The proposed \method demonstrates robust performance across both homophilic and heterophilic datasets, achieving the best average rank among all baselines. This indicates its effectiveness in handling diverse graph structures.
    \item The GNNs and methods like LP that use fixed low-pass filters generally do well on homophilic datasets but tend to underperform on heterophilic datasets.  Conversely, specialized models like GloGNN and LinkX, designed for heterophilic graphs, do not perform as well on homophilic datasets such as the ogbn-arxiv dataset.
    \item The GNNs equipped with learnable filters generally perform well on both types of datasets, as they can adapt their filters to the dataset's structural patterns. However, their performance is still not optimal. The proposed Node-MoE, which utilizes multiple ChebNetII as experts, significantly outperforms a single ChebNetII, especially on heterophilic datasets. This result validates the effectiveness of our node-wise filtering approach.
    \item We also compare the proposed \method with GMoE, which adapts the receptive field for each node but still applies traditional graph convolution with low-pass filters. We can find \method consistently outperforms GMoE across all datasets. 
\end{itemize}

\begin{table}[!htb]
\caption{Node classification accuracy (\%) on benchmark datasets. OOM  means out-of-memory. The bold and underline
markers denotes the best and second-best  performance respectively. }
\label{tab:performance}
\resizebox{\textwidth}{!}{%
\begin{tabular}{c|cccc|ccc|p{0.8cm}}
\toprule
 & \multicolumn{4}{c|}{Homophilic Datasets} & \multicolumn{3}{c|}{Heterophilc Datasets} & Avg. \newline Rank\\ \midrule
 & Cora & CiteSeer & PubMed & ogbn-arxiv & Chameleon & Squirrel & Actor & \makecell[c]{/} \\ \midrule
MLP & 76.49 ± 1.13 & 73.15 ± 1.36 & 86.14 ± 0.64 & 55.68 ± 0.11 & 48.11 ± 2.23 & 31.68 ± 1.90 & 36.17 ± 1.09 & \makecell[c]{10.71}\\
LP & 86.05 ± 1.35 & 69.39 ± 2.01 & 83.38 ± 0.64 & 68.14 ± 0.00 & 44.10 ± 4.10 & 31.92 ± 0.82 & 24.62 ± 0.51 & \makecell[c]{12}\\ \midrule
GCN & 88.60 ± 1.19 & 76.88 ± 1.78 & 88.48 ± 0.46 & 71.91 ± 0.15 & 67.96 ± 1.82 & 54.47 ± 1.17 & 30.31 ± 0.98 &\makecell[c]{6}\\
GAT & 88.68 ± 1.13 & 76.70 ± 1.81 & 86.52 ± 0.56 & 71.92 ± 0.17 & 65.29 ± 2.54 & 49.46 ± 1.69 & 28.95 ± 1.08 & \makecell[c]{7.29}\\
APPNP & 88.49 ± 1.28 & \underline{77.42 ± 1.47} & 87.56 ± 0.52 & 71.61 ± 0.30 & 54.32 ± 2.61 & 36.41 ± 1.94 & 36.05 ± 0.91 &\makecell[c]{7.43}\\
GCNII & 88.12± 1.05 & 77.30 ±1.58 & \textbf{90.17 ± 0.57} & \underline{72.74 ± 0.16} & 55.54 ± 2.02 & 56.63 ± 1.17 & 34.36 ± 0.77 &  \makecell[c]{5.43} \\ \midrule
GloGNN & \underline{88.78 ± 1.21} & 75.21 ± 1.17 & 87.93 ± 0.63 & 65.68 ± 0.04 & 69.78 ± 2.42 & 57.54 ± 1.39 & \textbf{37.35 ± 1.30} &\makecell[c]{5.43}\\
LinkX & 82.89 ± 1.27 & 70.05 ± 1.88 & 84.81 ± 0.65 & 66.54 ± 0.52 & 68.42 ± 1.38 & \underline{61.81 ± 1.80} & 36.10 ± 1.55 & \makecell[c]{8.14} \\
WRGCN &88.06 ± 1.50	&76.28 ± 1.98&	86.39 ± 0.55&	OOM &	65.24 ± 0.87 & 48.85 ± 0.78	&34.89 ± 1.11 & \makecell[c]{9.57}\\
\midrule
GPR-GNN & 88.54 ± 0.67 & 76.44 ± 1.89 & 88.46 ± 0.31 & 71.78 ± 0.18 & 62.85 ± 2.90 & 54.35 ± 0.87 & 33.12 ± 0.57 & \makecell[c]{7.29}\\
ChebNetII & 88.71 ± 0.93 & 76.93 ± 1.57 & 88.93 ± 0.29 & 72.32 ± 0.23 & 71.14 ± 2.13 & 57.12 ± 1.13 & 35.67 ± 1.19 & \makecell[c]{3.86} \\ \midrule
GMoE & 87.27 ± 1.74 & 76.56 ± 1.57 & 88.14 ± 0.56 & 71.74 ± 0.29 & \underline{71.88 ± 1.60} & 51.97 ± 3.16 & 35.75 ± 1.31 & \makecell[c]{6.57}\\ \midrule
\method & \textbf{89.38 ± 1.26} & \textbf{77.78 ± 1.36} & \underline{89.58 ± 0.60} & \textbf{73.19 ± 0.22} & \textbf{73.64 ± 1.80} & \textbf{62.31 ± 1.98} & \underline{36.28 ± 1.01} & \makecell[c]{1.29}\\ \bottomrule
\end{tabular}%
}
\end{table}

\subsection{Analysis of \method}
\label{sec:case_study}
In this section, we delve into an in-depth analysis of the behaviors of \method to demonstrate its rationality and effectiveness. We aim to uncover several key aspects of how \method operates and performs: What specific types of filters does Node-MoE learn? Are nodes appropriately assigned to these diverse filters by the gating model? 
We conduct experiments on both CiteSeer and Chameleon datasets using configurations with 2 experts. The results for the Chameleon dataset are presented below. For more results and analysis, please refer to Appendix~\ref{app:analysis}.

\begin{figure*}[htb]
\centering
\begin{minipage}[b]{.3\textwidth}
  \centering
  \includegraphics[height=4cm, keepaspectratio]{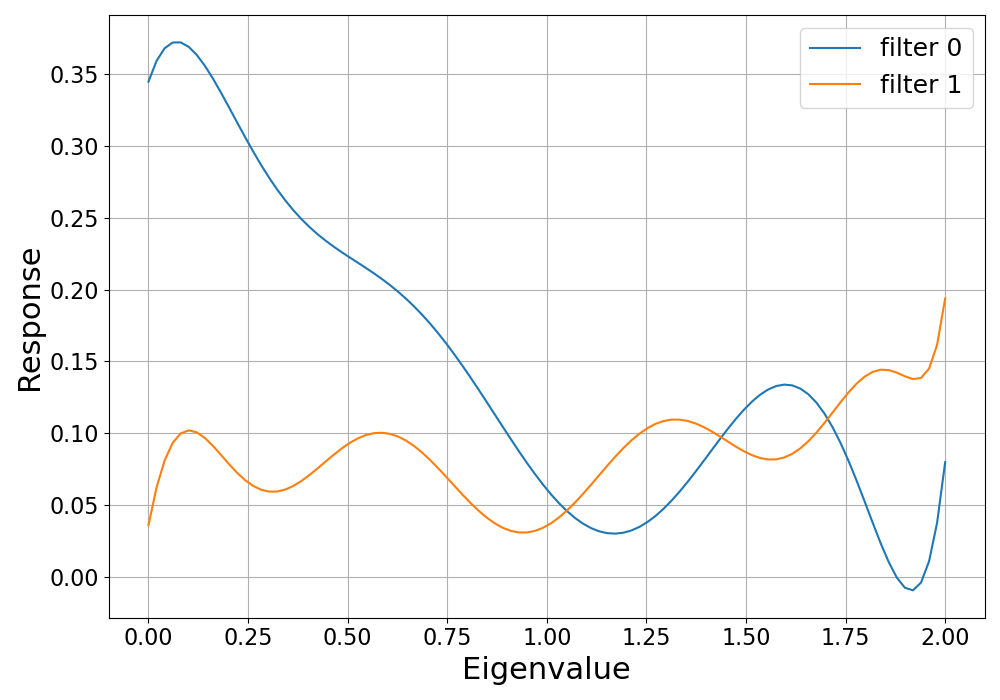}
\caption{Learned 2 filters by \method on Chameleon.}
\label{fig:filter2_chameleon}
\end{minipage} \hfill
\begin{minipage}[b]{.6\textwidth}
  \centering
  \includegraphics[height=4cm, keepaspectratio]{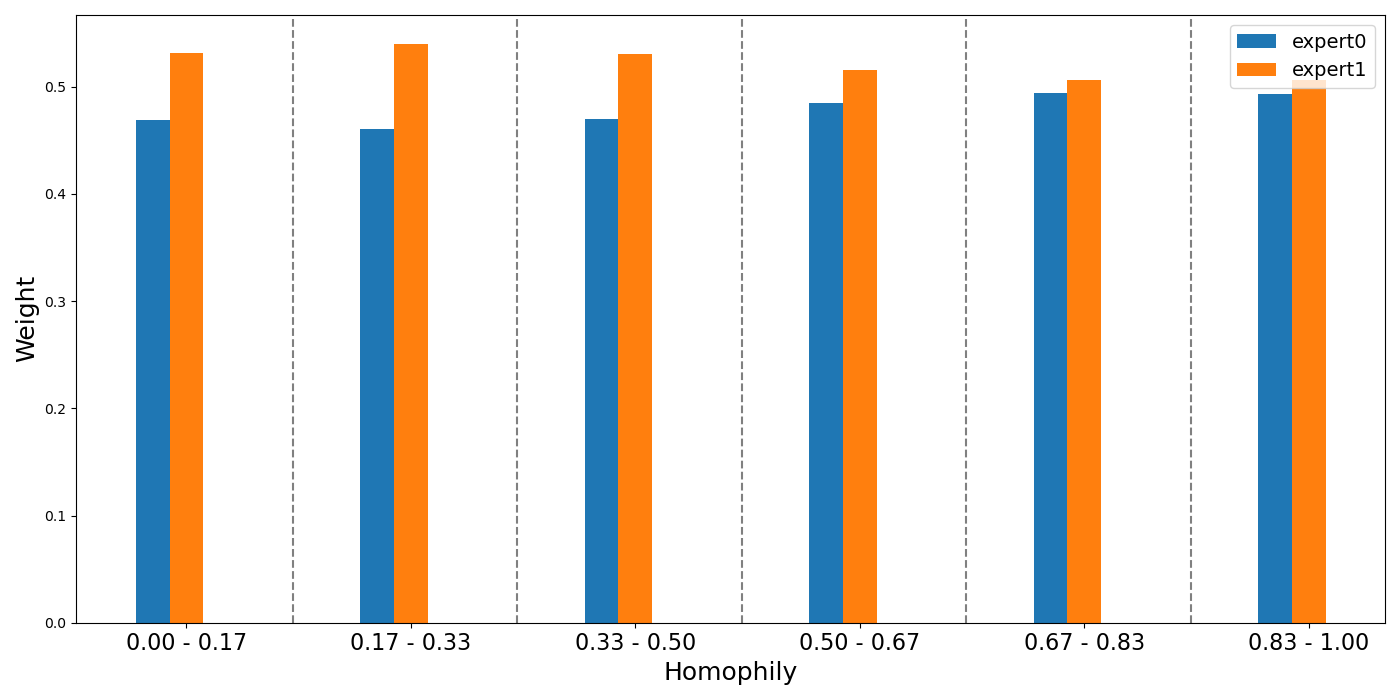}
    \caption{The average weight generated by the gating model for nodes in different homophily groups on Chameleon.}
 \label{fig:weight2_chameleon}
\end{minipage}
\end{figure*}

Figure~\ref{fig:filter2_chameleon} showcases the two filters learned by \method on the Chameleon dataset, where filter 0 functions as a low-pass filter and filter 1 as a high-pass filter. To analyze the behavior of the gating model in \method, we split nodes into different groups based on their homophily levels. Figure~\ref{fig:weight2_chameleon} displays the weights assigned by the gating model to these two experts. The results reveal that nodes with lower homophily levels predominantly receive higher weights for the high-pass filter (filter 1), and as the homophily level increases, the weight for this filter correspondingly decreases.  This pattern confirms our design that nodes with varying structural patterns require different filters, demonstrating the effectiveness of the proposed gating model.

\subsection{Ablation studies}
\label{sec:ablation}
\begin{figure*}[!htb]
\centering
\begin{minipage}[b]{.48\textwidth}
  \centering
  \includegraphics[width=1\linewidth, height=4cm, keepaspectratio]{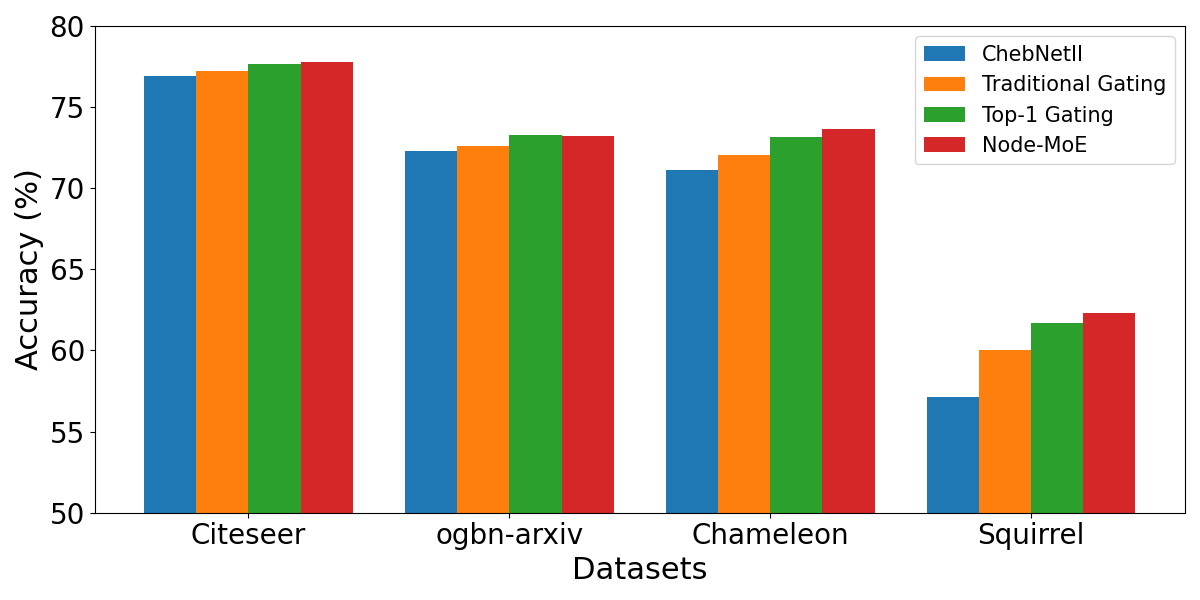}
\caption{The performance comparison of different gating variants.}
\label{fig:gating}
\end{minipage} \hfill
\begin{minipage}[b]{.48\textwidth}
  \centering
  \includegraphics[width=1\linewidth, height=3.4cm, keepaspectratio]{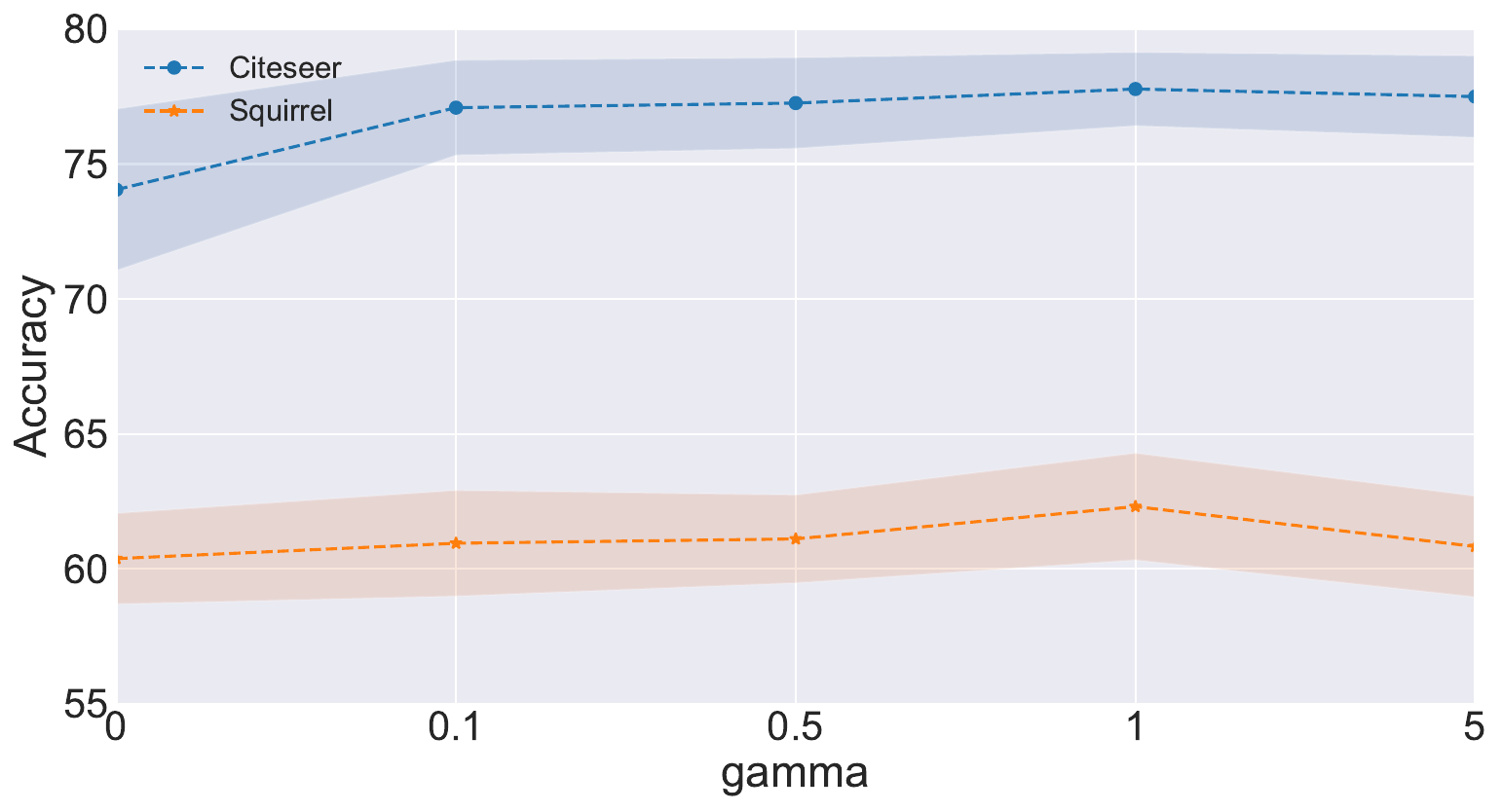}
    \caption{The performance with different weight parameters $\gamma$ of the filter smoothing loss.}
 \label{fig:gamma}
\end{minipage}
\end{figure*}

In this section, we conduct ablation studies to further investigate the effectiveness of two key components within the Node-MoE framework: the gating model and the filter smoothing loss. For the gating model, we explore two variants: a traditional MLP-based gating mechanism that utilizes the input features $\vX$, and the Top-k gating approach as detailed in Section~\ref{sec:topk}. Specifically, we choose $k=1$ to ensure the proposed \method has similar efficiency with the single expert.
Figure~\ref{fig:gating} presents the results on CiteSeer, ogbn-arxiv, Chameleon, and Squirrel datasets. We observe two findings: (1) Traditional gating does not perform as well as the proposed gating in \method and only achieves results comparable to an individual ChebNetII expert. (2) The Top-1 gating, which selects only one expert, can achieve similar results to those of the soft gating \method that utilizes all experts.  This indicates that the proposed \method can effectively enhance performance while maintaining a complexity level comparable to that of an individual expert model.

We also investigate the impact of the weight parameter, $\gamma$, of the filter smoothing loss on the overall performance. Specifically, we conduct experiments on the Citeseer and Squirrel datasets and the $\gamma$ is chosen in $[0, 0.1, 0.5, 1, 5]$. As shown in Figure~\ref{fig:gamma}, incorporating the filter smoothing loss generally enhances performance, especially for the Citeseer dataset. The reason is that the filter smoothing loss can mitigate filter oscillation, which may lead to the model being hard to learn.  For more detailed insights into the effects of the filter smoothing loss, please refer to Appendix~\ref{app:smoothing}.

\section{Related Works}
\label{sec:relate}
Graph Neural Networks (GNNs)~\citep{kipf2016semi, velivckovic2017graph}  have achieved remarkable success in graph representation learning across a wide range of tasks~\citep{zhou2020graph}. The design of GNN architectures is majorly motivated in the spatial domain~\citep{kipf2016semi, hamilton2017inductive, gasteiger2018predict, gasteiger2019diffusion} and spectral domain~\citep{defferrard2016convolutional, chien2020adaptive, levie2018cayleynets, bianchi2021graph}. The GNNs in the spatial domain usually follow the message-passing mechanism~\citep{gilmer2017neural}, which can be regarded as low-pass graph filters~\citep{nt2019revisiting, zhao2019pairnorm}. As a result, these GNNs are usually suitable for homophilic graphs.  To address heterophilic graphs, specialized models like GloGNN~\citep{li2022finding}, LinkX~\citep{lim2021large}, MixHop~\cite{abu2019mixhop}, and ACM-GNN~\cite{luan2022revisiting} have been developed. Additionally, models such as Bernnet~\citep{he2021bernnet}, GPRGNN~\citep{chien2020adaptive}, and ChebNetII~\citep{he2022convolutional} feature learnable filters that adapt to various graph types. Recent studies have highlighted that real-world graphs often exhibit a mixture of structural patterns~\citep{li2022finding, mao2024demystifying}, with \cite{mao2024demystifying} specifically exploring GNN generalization under mixed patterns.  Traditional GNNs typically apply the same global filter across all nodes, which can be suboptimal for such mixed scenarios. In response, our proposed \method introduces a node-wise filtering approach, applying distinct filters to nodes based on their individual patterns, enhancing adaptability and performance.

Mixture of Experts (MoE)~\citep{jacobs1991adaptive, jordan1994hierarchical} architecture has been widely used in NLP~\citep{du2022glam, zhou2022mixture} and Computer Vision~\citep{riquelme2021scaling} to improve efficiency of large models. In graph domain, GraphMETRO~\citep{wu2023graphmetro} leverage
MoE to address the graph distribution shift issue. GMoE~\citep{wang2024graph} utilizes MoE to adaptive select propagation hops for different nodes. Despite these advancements, these method still face challenges in handling complex graph patterns with the same efficacy as our proposed \method.

\vspace{-0.1in}
\section{Conclusion}
In this paper, we explored the complex structural patterns inherent in real-world graph datasets, which typically showcase a mixture of homophilic and heterophilic patterns. Notably, these patterns exhibit significant variability across different communities within the same dataset, highlighting the intricate and diverse nature of graph structures. Our theoretical analysis reveals that the conventional single global filter, commonly used in many GNNs, is often inadequate for capturing such complex structural patterns. To address this limitation, we proposed the Node-wise filtering method, \method, a flexible and effective solution that adaptively selects appropriate filters for different nodes. Extensive experiments demonstrate the proposed \method demonstrated robust performance on both homophilic and heterophilic datasets.  Further, our behavioral analysis and ablation studies validate the design and effectiveness of the proposed \method.

\bibliographystyle{unsrt}
\bibliography{reference}

\newpage
\appendix
\renewcommand \thepart{} 
\renewcommand \partname{}
\part{Appendix}

\section{Proof of Theorem 1}
\label{app:proof}
In this section, we present the proof of Theorem~\ref{theorem1}. This theorem analyzes the separability when different filters are applied to graphs generated by a mixed CSBM model in Defination~\ref{def1}-$CSBM(n, \bmu, \bnu, (p_0, q_0), (p_1, q_1), P)$ using a linear classifier. 

\textbf{Notably, the following proof is derived based on \cite{baranwal2021graph}}, which analyzes the linear separability of a single graph convolution under a single CSBM model with only one pattern - $CSBM(n, \bmu, \bnu, (p, q))$. We extend the analysis to graphs with mixed CSBM models. Besides, we analyze the scenarios in which different filters are applied to the same graph.

We follow the assumption 1 and 2 in \cite{baranwal2021graph}: The graph size n should be relatively large with $\omega(d \log d) \leq n \leq O(\operatorname{poly}(d))$, and the graph is not too sparse with $p_0, q_0, p_1, q_1=\omega\left(\log ^2(n) / n\right)$.

\subsection{Proof of part 1 of Theorem 1}
\begin{proof}
For the low-pass filter, consider the filtered feature $\tilde{\vX} = \vD^{-1}\vA\vX$. Then, the filtered feature of node $i$ still follows the normal distribution, the mean can be represented by:

$
m(i) = E(\tilde{\vX}_i) = \left\{
\begin{aligned}
 \frac{p_0\bmu+q_0\bnu}{p_0 + q_0} (1+o(1)) \quad \text{for}  \, i \in C_0  \, \text{and}  \, i \in H_0 \\
 \frac{q_0\bmu+p_0\bnu}{p_0 + q_0} (1+o(1)) \quad \text{for}  \, i \in C_1  \, \text{and}  \, i \in H_0 \\
 \frac{p_1\bmu+q_1\bnu}{p_1 + q_1} (1+o(1)) \quad \text{for}  \, i \in C_0  \, \text{and}  \, i \in H_1 \\
  \frac{q_1\bmu+p_1\bnu}{p_1 + q_1} (1+o(1)) \quad \text{for}  \, i \in C_1  \, \text{and}  \, i \in H_1 \\
\end{aligned}
\right.,
$

where $C_0$ and $C_1$ represent the class 0 and class 1, respectively; $H_0$ and $H_1$ are the homophilic and heterophilic node sets, respectively. The covariance matrix can be represented by:
$
Cov(\tilde{\vX}_i) = \frac{1}{d\vD_{ii}} \vI.
$. Lemma 2 in \cite{baranwal2021graph} demostrate that for any unit vector $\vw$, we have:
$
\left|\left(\tilde{\vX}_i-m(i)\right) \cdot \mathbf{w}\right|=O\left(\sqrt{\frac{\log n}{d n(p_0+q_0)}}\right).
$

If we only consider the nodes with homophilic patterns, i.e., $i \in H_0$, we can find the optimal linear classifier with $\vw^* = R\frac{\bnu-\bmu}{\|\bnu - \bmu\|}$ and $\vb^* = -\frac{1}{2} \langle \bnu + \bmu, \vw^* \rangle$. We also have the assumption that the distance between $\bmu$ and $\bnu$ are relatively large, with $\|\bnu - \bmu\| = \Omega\left(\frac{\log n}{d n(p_0+q_0)}\right)$.

Then, for $i \in C_0  \, \text{and}  \, i \in H_0 $, we have:

\begin{align*}
    \langle \tilde{\vX}_i, \vw^* \rangle + b^* &= \frac{\langle p_0\bmu+q_0\bnu, \vw^* \rangle}{p_0 + q_0} (1+o(1)) + O\left(\|\vw^*\| \sqrt{\frac{\log n}{d n(p+q)}}\right) - \frac{1}{2} \langle \bnu + \bmu, \vw^* \rangle \\
    &= \frac{\langle 2p_0\bmu + 2q_o\bnu - (p_0 + q_0)(\bmu+\bnu), \vw^*\rangle}{p_0 + q_0}(1+o(1)) + o(\|\vw^*\|) \\
    &= \frac{p_0-q_0}{2(p_0+q_0)} \langle \bmu - \bnu, \vw*\rangle (1+o(1)) + o(\|\vw^*\|)\\
    &= -\frac{R(p_0-q_0)}{2(p_0+q_0)}\|\bmu - \bnu\|(1+o(1)) < 0
\end{align*}

Similarly, for $i \in C_1  \, \text{and}  \, i \in H_0 $, we have:
\begin{align*}
    \langle \tilde{\vX}_i, \vw^* \rangle + b^*
    &= -\frac{R(q_0-p_0)}{2(p_0+q_0)}\|\bmu - \bnu\|(1+o(1)) > 0
\end{align*}

Therefore, the linear classifier with $w^*$ and $b^*$ can separate class $C_0$. 

However, if we apply this linear classifier to the heterophilic node set $H_1$, we have:

$
\langle \tilde{\vX}_i, \vw^* \rangle + b^* = \left\{
\begin{aligned}
-\frac{R(p_1-q_1)}{2(p_1+q_1)}\|\bmu - \bnu\|(1+o(1)) > 0 \quad \text{for}  \, i \in C_0  \, \text{and}  \, i \in H_1 \\
-\frac{R(q_1-p_1)}{2(p_1+q_1)}\|\bmu - \bnu\|(1+o(1)) < 0 \quad \text{for}  \, i \in C_1  \, \text{and}  \, i \in H_1 \\
\end{aligned}
\right.
$

Therefore, all nodes in $H_1$ are misclassified. The binary cross-entropy over node set $H_1$ can be represented by:
\begin{align*}
L(H_1, \vw^*, b^*) & =\frac{1}{|H_1|}\sum_{i \in H_1} -y_i \log \left(\sigma\left(\left\langle\tilde{\vX}_i, \vw^* \right\rangle+\tilde{b}\right)\right)-\left(1-y_i\right) \log \left(1-\sigma\left(\left\langle\tilde{\vX}_i, \vw^*\right\rangle+b^*\right)\right) \\
& =\frac{1}{|H_1|}\sum_{i \in H_1} \log \left(1+\exp \left(\left(1-2y_i \right)\left(\left\langle\tilde{X}_i, \tilde{\mathbf{w}}\right\rangle+b^*\right)\right)\right) \\
&=\log \left(1+\exp \left( -\frac{R(p_1-q_1)}{2(p_1+q_1)}\|\bmu - \bnu\|(1+o(1))  \right)\right)
\end{align*}

As for $x = -\frac{R(p_1-q_1)}{2(p_1+q_1)}\|\bmu - \bnu\| > 0$, we have $e^x \geq x$. As a result, we have
\begin{align*}
    L(H_1, \vw^*, b^*) \geq \frac{R(q_1-p_1)}{2(p_1+q_1)}\|\bmu - \bnu\|(1+o(1))
\end{align*}

\end{proof}

\subsection{Proof of part 2 of Theorem 1}
\begin{proof}
    Suppose we apply a high-pass filter to the heterophilic nodes $H_1$ and the filtered features are $\tilde{\vX} = -\vD^{-1}\vA\vX$. For nodes in $H_1$, 
    
    $
m(i) = E(\tilde{\vX}_i) = \left\{
\begin{aligned}
 -\frac{p_1\bmu+q_1\bnu}{p_1 + q_1} (1+o(1)) \quad \text{for}  \, i \in C_0  \, \text{and}  \, i \in H_1 \\
 - \frac{q_1\bmu+p_1\bnu}{p_1 + q_1} (1+o(1)) \quad \text{for}  \, i \in C_1  \, \text{and}  \, i \in H_1 \\
\end{aligned}
\right.
$

Therefore, if we apply the same linear classifier with $\vw^*$ and $b^*$, then we have:

$
\langle \tilde{\vX}_i, \vw^* \rangle + b^* = \left\{
\begin{aligned}
\frac{R(p_1-q_1)}{2(p_1+q_1)}\|\bmu - \bnu\|(1+o(1)) < 0 \quad \text{for}  \, i \in C_0  \, \text{and}  \, i \in H_1 \\
\frac{R(q_1-p_1)}{2(p_1+q_1)}\|\bmu - \bnu\|(1+o(1)) > 0 \quad \text{for}  \, i \in C_1  \, \text{and}  \, i \in H_1 \\
\end{aligned}
\right.
$

As a result, the same linear classifier can separate both the homophilic set $H_0$ and heterophilic set $H_1$.

\end{proof}

\section{The Impact of Filter Smoothing Loss}
\label{app:smoothing}

\begin{figure*}[htb]
\centering
\begin{minipage}[b]{.48\textwidth}
  \centering
  \includegraphics[height=4cm, keepaspectratio]{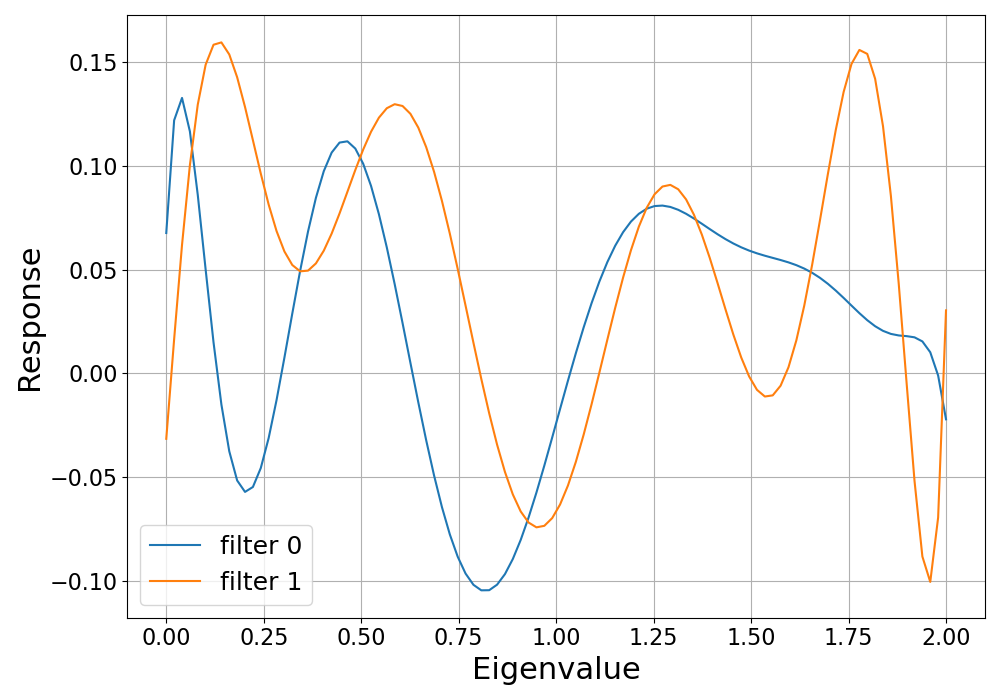}
\caption{Learned 2 filters by \method on Chameleon without filter smoothing loss.}
\label{fig:filter2_chameleon_0}
\end{minipage} \hfill
\begin{minipage}[b]{.48\textwidth}
  \centering
  \includegraphics[height=4cm, keepaspectratio]{Figures/chameleon_MoE_filter_2_1.png}
    \caption{Learned 2 filters by \method on Chameleon with filter smoothing loss.}
 \label{fig:filter2_chameleon_1}
\end{minipage}
\end{figure*}

In this section, we explore the impact of the proposed filter smoothing loss on the behavior of the learned filters in our \method framework. Figures~\ref{fig:filter2_chameleon_0} and \ref{fig:filter2_chameleon_1} display the effects of the \method framework without and with the application of filter smoothing loss, respectively. Without the filter smoothing loss, as shown in Figure~\ref{fig:filter2_chameleon_0}, the learned filters exhibit significant oscillations, making it challenging to discern their specific functions. In contrast, with the filter smoothing loss applied, as illustrated in Figure~\ref{fig:filter2_chameleon_1}, the behavior of the filters becomes more distinct: filter 0 clearly functions as a low-pass filter, and filter 1 as a high-pass filter. 

Additionally, we assessed the training dynamics of the proposed Node-MoE framework by comparing performance with and without the filter smoothing loss, while keeping other hyperparameters constant. For the Citeseer dataset, applying the filter smoothing loss resulted in a higher average training accuracy of 99.37 ± 0.17, compared to 93.51 ± 1.27 when the loss was not applied. A similar pattern was observed on the Squirrel dataset, where the training accuracy was 96.54 ± 1.42 with the filter smoothing loss, versus 95.54 ± 0.94 without it. These results suggest that oscillations in the filters without the smoothing loss can hinder the model’s ability to fit the data effectively, resulting in suboptimal performance as shown in Section~\ref{sec:ablation}.

\section{Datasets and Experimental Settings}
\label{app:setting}
In this section, we detail the datasets used and the experimental settings for both the baseline models and the proposed \method framework. 

\subsection{Datasets}
\label{sec:app_data}

We conduct experiments across seven widely recognized datasets, which encompass both homophilic and heterophilic types. The homophilic datasets include Cora, CiteSeer, and Pubmed~\citep{sen2008collective}, along with ogbn-arxiv~\citep{hu2020open}; the heterophilic datasets comprise Chameleon, Squirrel, and Actor~\citep{pei2020geom}. For Cora, CiteSeer, and Pubmed, we generate ten random splits, allocating nodes into training, validation, and testing sets with proportions of 60\%, 20\%, and 20\%, respectively. For the heterophilic datasets, we adhere to the ten fixed splits as defined in \cite{pei2020geom}. The ogbn-arxiv dataset is assessed using its standard split as established by \citep{hu2020open}. Detailed statistics of these datasets are shown in Table~\ref{table:app_data}.

\begin{table}[h]
\centering

 \caption{Statistics of datasets. The split ratio is for train/validation/test. 
 }
  \begin{adjustbox}{width =1 \textwidth}
\begin{tabular}{c|cccc|ccc}
\toprule
     & \multicolumn{4}{c|}{Homophilic Datasets} & \multicolumn{3}{c}{Heterophilc Datasets} \\ 
 & Cora & CiteSeer & PubMed & ogbn-arxiv   & Chameleon & Squirrel & Actor \\
 \midrule
\#Nodes & 2,708 & 3,327 & 19,717 & 169, 343 & 2,277 &  5,201 &  7,600\\
\#Edges & 5,429 & 4,732& 44,338 & 1, 166, 243 & 31,421 &198,493 &26,752\\
\#Classes & 7 &6 & 3 & 40 & 5 &5 &5\\
\#Node Features & 1,433 &3,703 &500 &128 & 2,325 &2,089 &931 \\
\#Split Ratio & 60/20/20 & 60/20/20 & 60/20/20 &  54/18/28 &60/20/20 & 60/20/20 & 60/20/20 \\
\bottomrule
\end{tabular}
\label{table:app_data}
\end{adjustbox}
\end{table}

\subsection{Experimental Settings}
\label{sec:app_expsetting}
For the baseline models, we adopt the same parameter setting in their original paper. For the proposed \method, we adopt the GIN as gating and GCNII as experts. Notably, the GCNII model has different learning rates and weight decay for the filters and other parameters. All the hyperparameters are tuned based on the validation accuracy from the following search space:
\begin{itemize}
    \item Gating Learning Rate: \{0.0001, 0.001, 0.01 \}
    \item Gating Dropout: \{0, 0.5, 0.8\}
    \item Gating Weight Decay: \{0, 5e-5, 5e-4\}
    \item Expert Learning Rate for Filters: \{0.001, 0.01, 0.1\}
    \item Expert Weight Decay for Filters: \{0, 5e-5, 5e-3, 5e-2 \}
    \item Expert Learning Rate: \{0.001, 0.01, 0.1, 0.5\}
    \item Expert Dropout:  \{0, 0.5, 0.8\}
    \item Filter Smoothing loss weight: \{0, 0.01, 0.1, 1\}
    \item Load balancing weight for top-k gating: \{0, 0.001, 0.01, 0.1, 1\}
    \item Number of experts: \{2, 3, 5\}
\end{itemize}

For the initialization of filters in ChebNetII, which uses a K-order approximation, we employ a set of initialization strategies for the polynomial coefficients. These strategies include: decreasing powers $[\alpha^0, \alpha^1, \cdots, \alpha^K]$, increasing powers $[\alpha^K, \alpha^{K-1}, \cdots, \alpha^0]$, and uniform values $[1, 1, \cdots, 1]$. For configurations with 2 or 3 experts, we set $\alpha = 0.9$. When expanding to 5 experts, we use two values of $\alpha$, setting them at $0.9$ and $0.8$, respectively, to diversify the response characteristics of the filters.

We use a single GPU of  NVIDIA RTX A5000 24Gb, to run the experiments.

\section{Analysis of the proposed \method}
\label{app:analysis}
In this section, we provide more analysis of the proposed \method by comprehensive experiments.

\subsection{The behavior of \method with 2 experts}

\begin{figure*}[htb]
\centering
\begin{minipage}[b]{.3\textwidth}
  \centering
  \includegraphics[height=4cm, keepaspectratio]{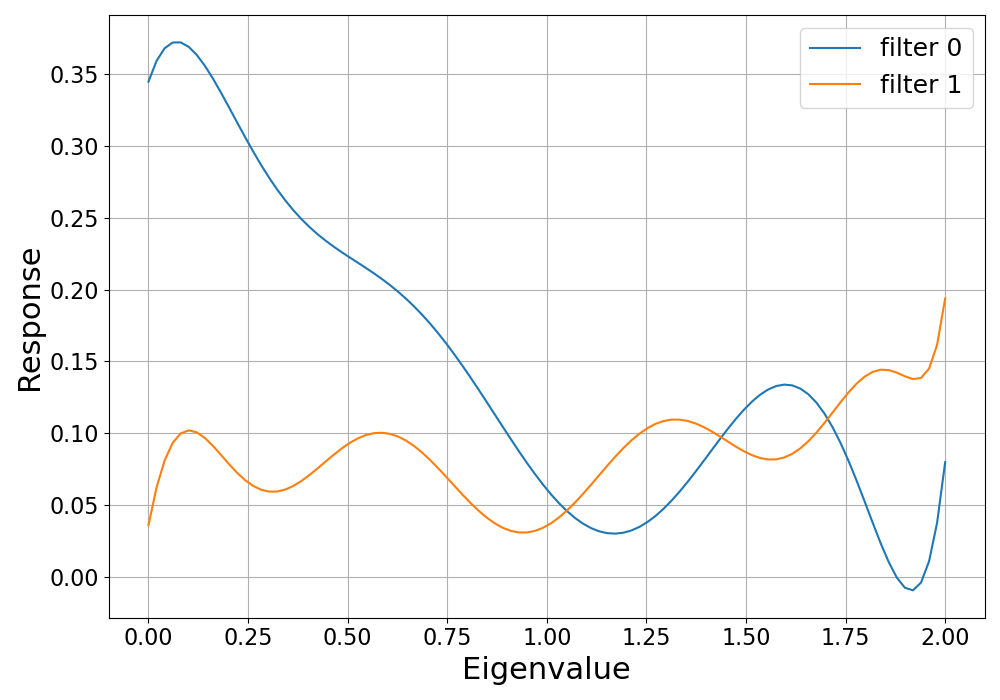}
\caption{Learned 2 filters by \method on Chameleon.}
\label{fig:filter2_chameleon_1_1}
\end{minipage} \hfill
\begin{minipage}[b]{.6\textwidth}
  \centering
  \includegraphics[height=4cm, keepaspectratio]{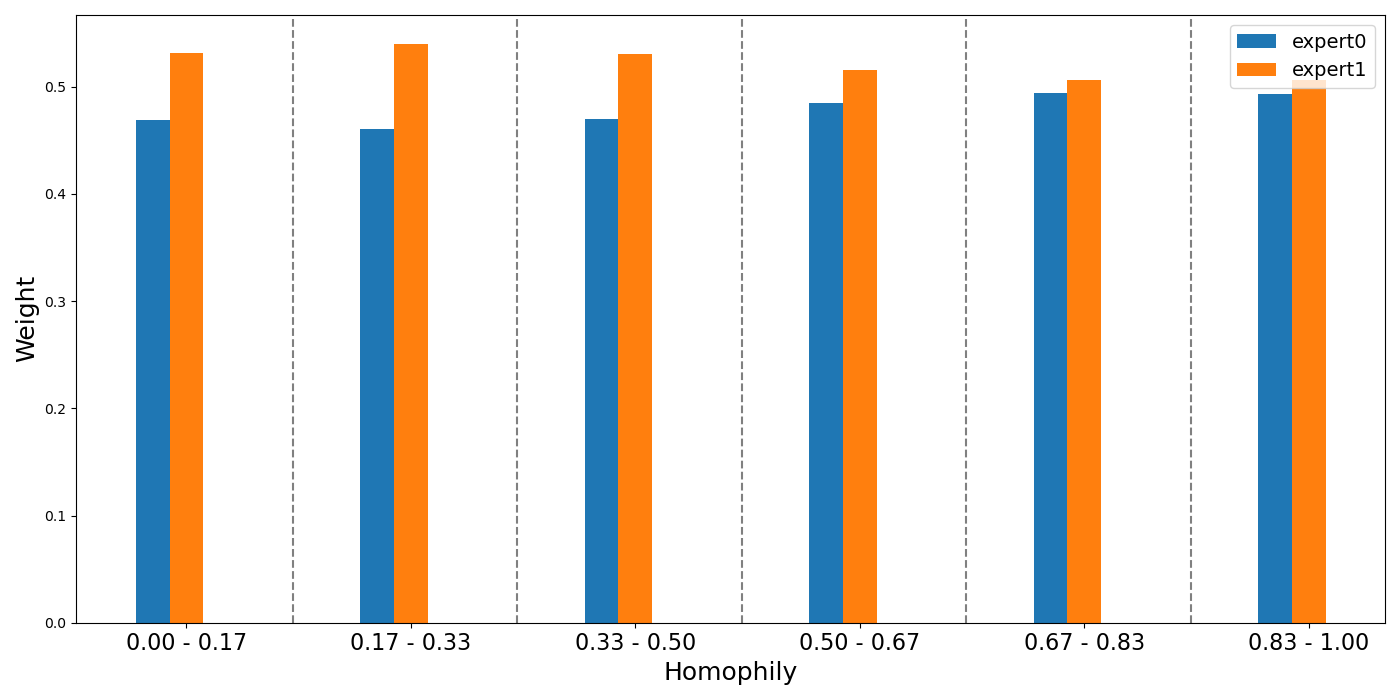}
    \caption{The average weight generated by the gating model for nodes in different homophily groups on Chameleon.}
 \label{fig:weight2_chameleon_1}
\end{minipage}
\end{figure*}

\begin{figure*}[htb]
\centering
\begin{minipage}[b]{.3\textwidth}
  \centering
  \includegraphics[height=4cm, keepaspectratio]{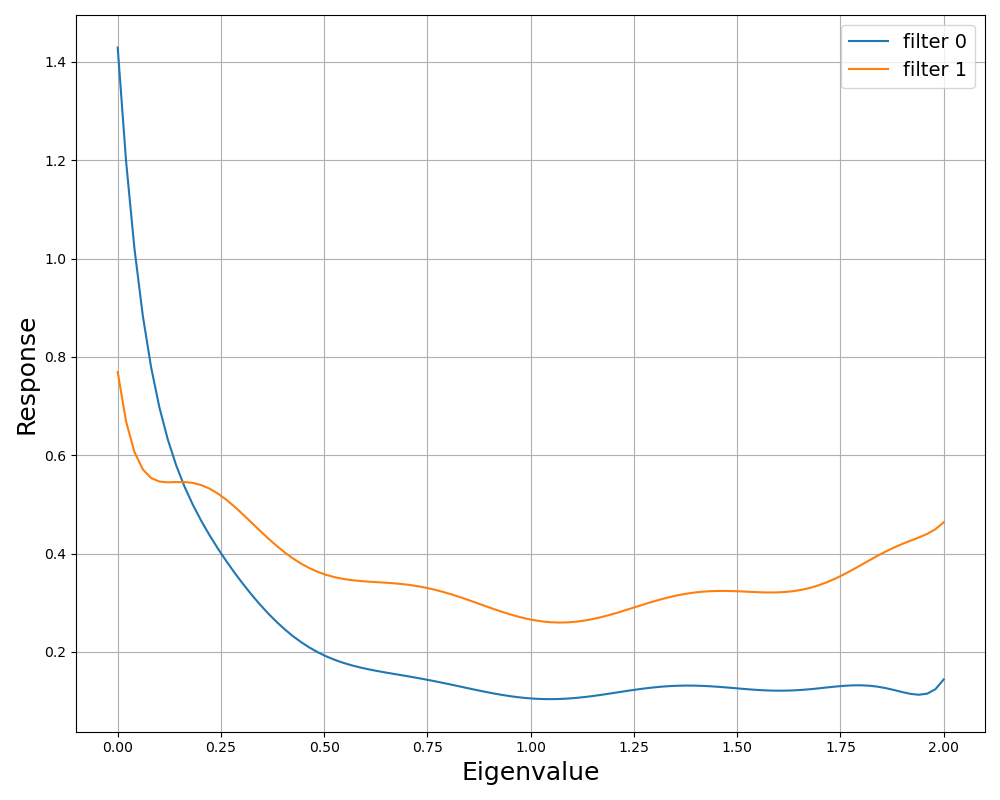}
\caption{Learned 2 filters by \method on Citeseer.}
\label{fig:filter2_citeseer_1}
\end{minipage} \hfill
\begin{minipage}[b]{.6\textwidth}
  \centering
  \includegraphics[height=4cm, keepaspectratio]{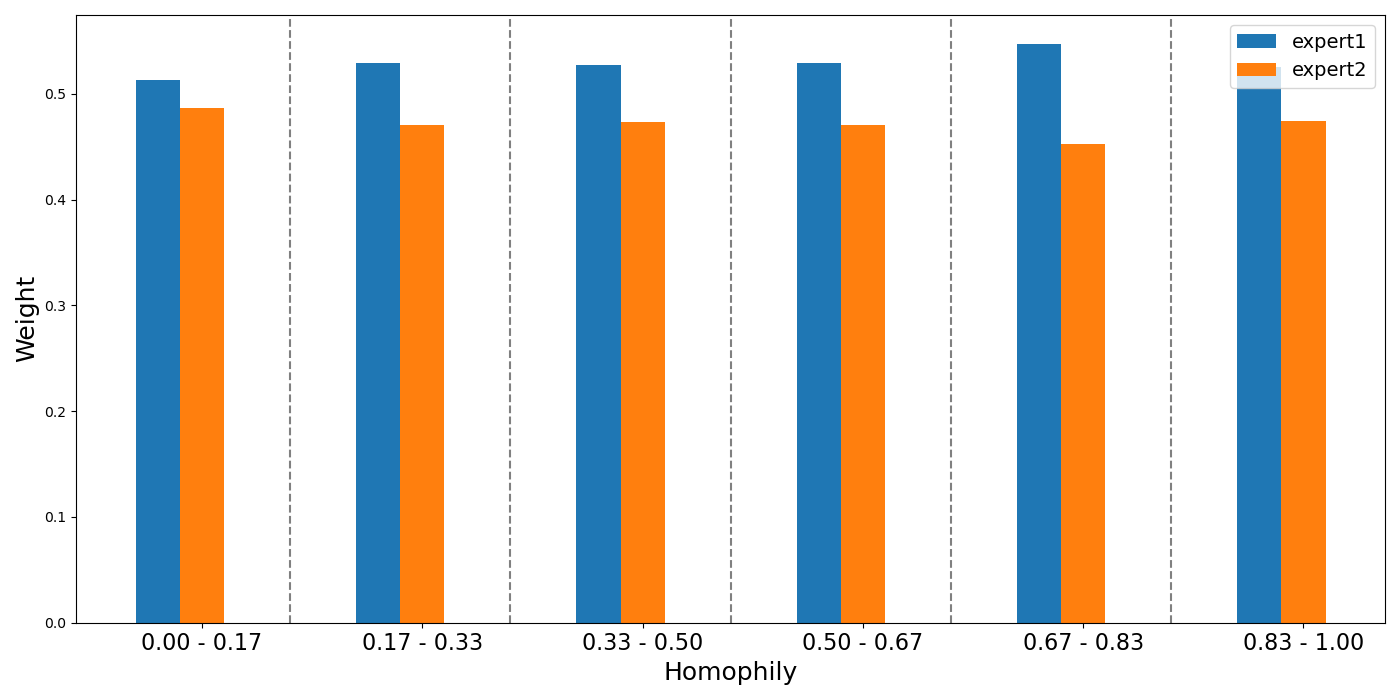}
    \caption{The average weight generated by the gating model for nodes in different homophily groups on Citeseer.}
 \label{fig:weight2_citeseer_1}
\end{minipage}
\end{figure*}

The learned filters and the corresponding gating weights for nodes with different homophily levels are illustrated below. For the Chameleon dataset, these are displayed in Figure~\ref{fig:filter2_chameleon_1_1} for the filters and Figure~\ref{fig:weight2_chameleon_1} for the gating weights. Similarly, for the Citeseer dataset, the filters are shown in Figure~\ref{fig:filter2_citeseer_1} and the gating weights in Figure~\ref{fig:weight2_citeseer_1}.

For both datasets, the learned filters demonstrate distinct characteristics: filter 0s function as low-pass filters, effectively smoothing signals, while filter 1s respond more strongly to high-frequency signals, characteristic of high-pass filters. Specifically, for the heterophilic dataset, such as Chameleon, the gating model generally assigns higher weights to filter 1, indicating a preference for high-pass filtering to accommodate the less homophilic nature of the dataset. Conversely, for the homophilic dataset, such as Citeseer, higher weights are typically assigned to filter 0, emphasizing low-pass filtering.

Moreover, within the Chameleon dataset, the weight assigned to the high-pass filter (filter 1) decreases as the homophily level increases. In contrast, in the Citeseer dataset, the weight to the low-pass filter (filter 0) increases with rising homophily levels. This pattern supports our initial hypothesis: nodes with lower homophily are better served by high-pass filters to capture the dissimilarity among neighbors, while nodes with higher homophily benefit from low-pass filters to reinforce the similarity among neighboring nodes.

\subsection{Accuracy analysis between the \method with the individual expert}
In this section, we evaluate the performance differences between the proposed Node-MoE framework and a single expert model, specifically ChebNetII, across nodes with varying homophily levels in the Chameleon dataset. 

\begin{figure*}[htb]
\centering
\begin{minipage}[b]{.3\textwidth}
  \centering
  \includegraphics[height=4cm, keepaspectratio]{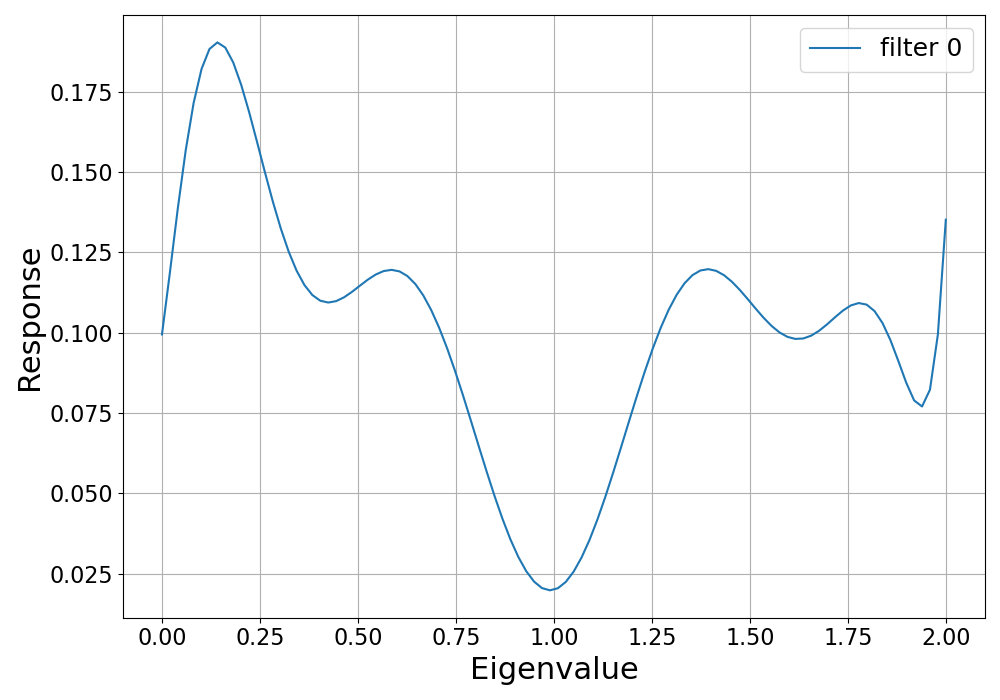}
\caption{The filter learned by ChebNetII on Chameleon.}
\label{fig:filter1_chameleon}
\end{minipage} \hfill
\begin{minipage}[b]{.6\textwidth}
  \centering
  \includegraphics[height=4cm, keepaspectratio]{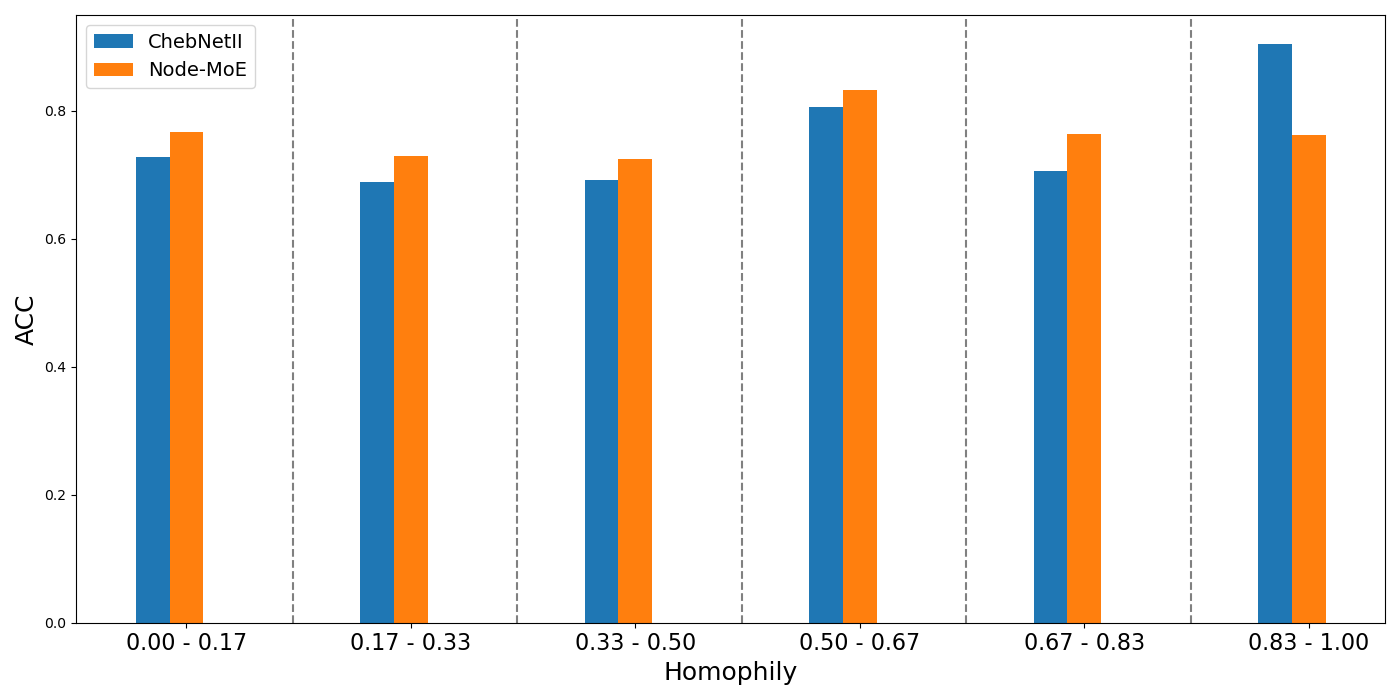}
    \caption{The average accuracy for nodes in different homophily groups on Chameleon.}
 \label{fig:acc_chameleon}
\end{minipage}
\end{figure*}

The filter characteristics learned by the individual ChebNetII are depicted in Figure~\ref{fig:filter1_chameleon}, which shows that this filter responds strongly to both low and high-frequency components. The comparative performance analysis is presented in Figure~\ref{fig:acc_chameleon}. From these results, it is evident that ChebNetII only outperforms Node-MoE when the node homophily level is exceptionally high. These findings indicate that the reliance on a single global filter, such as with ChebNetII, can lead to suboptimal performance. While such a filter might fit exceptionally well for nodes of a particular type, it may not adequately address the needs of nodes in other groups, thereby limiting overall effectiveness across diverse node types.

\end{document}